\documentclass[11pt]{article} 
\usepackage{rldmsubmit,palatino}
\usepackage{amsmath,amsfonts,amsthm}
\usepackage{graphicx}
\usepackage{dsfont}
\usepackage{mathtools}
\usepackage{tikz}
\usetikzlibrary{arrows} 
\usetikzlibrary{math}
\usepackage{pgfplots}
\usepackage{thm-restate}
\usepackage{algorithm}
\usepackage{algpseudocode}
\usepackage{todonotes}
\usepackage{hyperref}
\usepackage{wrapfig}
\usepackage{verbatim}
\usepackage{multirow}

\usepackage{ulem}

\tikzset{
>=stealth',
punkt/.style={
           rectangle,
           rounded corners,
           draw=black, very thick,
           text width=6.5em,
           minimum height=2em,
           text centered},
pil/.style={
           ->,
           thick,
           shorten <=2pt,
           shorten >=2pt,}
}
\pgfplotsset{
  grid style={gray, dashed, very thin},
  every inner x axis line/.append style={pil},
  every inner y axis line/.append style={pil},
}

\theoremstyle{definition}
\newtheorem{definition}{Definition}[]

\newtheorem{assumption}[definition]{Assumption}
\newtheorem*{goal*}{Goal}
\newtheorem*{setting*}{Setting}
\newtheorem*{assumption*}{Assumption}
\theoremstyle{plain}

\renewcommand{\AA}{\mathcal{A}}
\renewcommand{\SS}{\mathcal{S}}
\newcommand{\OO}{\mathcal{O}}

\title{Solving infinite-horizon POMDPs with memoryless\\stochastic policies in state-action space   
}

\author{
Johannes Müller 
\\
Max Planck Institute for Mathematics in the Sciences, Leipzig, Germany \\
\texttt{jmueller@mis.mpg.de} \\
\And
Guido Mont\'ufar\\
Department of Mathematics and Department of Statistics, UCLA, CA, USA \\
Max Planck Institute for Mathematics in the Sciences, Leipzig, Germany \\
\texttt{montufar@math.ucla.edu}\\
}

\begin{document}

\maketitle

\begin{abstract}
Reward optimization in fully observable Markov decision processes is equivalent to a linear program over the polytope of state-action frequencies. 
Taking a similar perspective in 
the case of partially observable Markov decision processes with memoryless stochastic policies, the problem was recently formulated as the optimization of a linear objective subject to polynomial constraints. 
Based on this we present an approach for Reward Optimization in State-Action space (ROSA). 
We test this approach experimentally in maze navigation tasks. We find that ROSA is computationally efficient and can yield stability improvements over other existing methods. 
\end{abstract} 


\acknowledgements{
The authors acknowledge support from the ERC under the European Union’s Horizon 2020 research and innovation programme (grant agreement no 757983). 
JM also acknowledges support from 
the International Max Planck Research School for Mathematics in the Sciences and the Evangelisches Studienwerk Villigst e.V..}  

\startmain 

\normalem
\section{Introduction}

Partially Observable Markov Decision Processes (POMDPs) offer a popular model for sequential decision making with state uncertainty. Here, actions are selected based on partial observations of the system's state with the objective to maximize a cumulative discounted reward. 
We focus on infinite horizon problems and  
memoryless stochastic policies which offer an alternative to 
difficult-to-optimize 
policies based on belief states and policies with memory. 
%
Techniques such as policy iteration and value iteration 
usually require a belief state or a finite-state controller representation of the policy, and the standard approaches based on policy gradients 
\cite{sutton1999policy,azizzadenesheli2018policy} can suffer from ill-conditioning when future rewards are not sufficiently discounted. 
{
Recently, a polynomial programming formulation of POMDPs was derived in~\cite{mueller2021geometry}, generalizing the linear program associated to MDPs.} 
{
In this work we provide a practical implementation of this approach and demonstrate experimentally, in navigation problems of different sizes, that it offers a competitive alternative improving computational cost and numerical stability for a range of discount factors. 
}

\begin{wrapfigure}{r}{9cm}
    \centering
    \vspace{-.5cm}
    \includegraphics[width=0.28\textwidth]{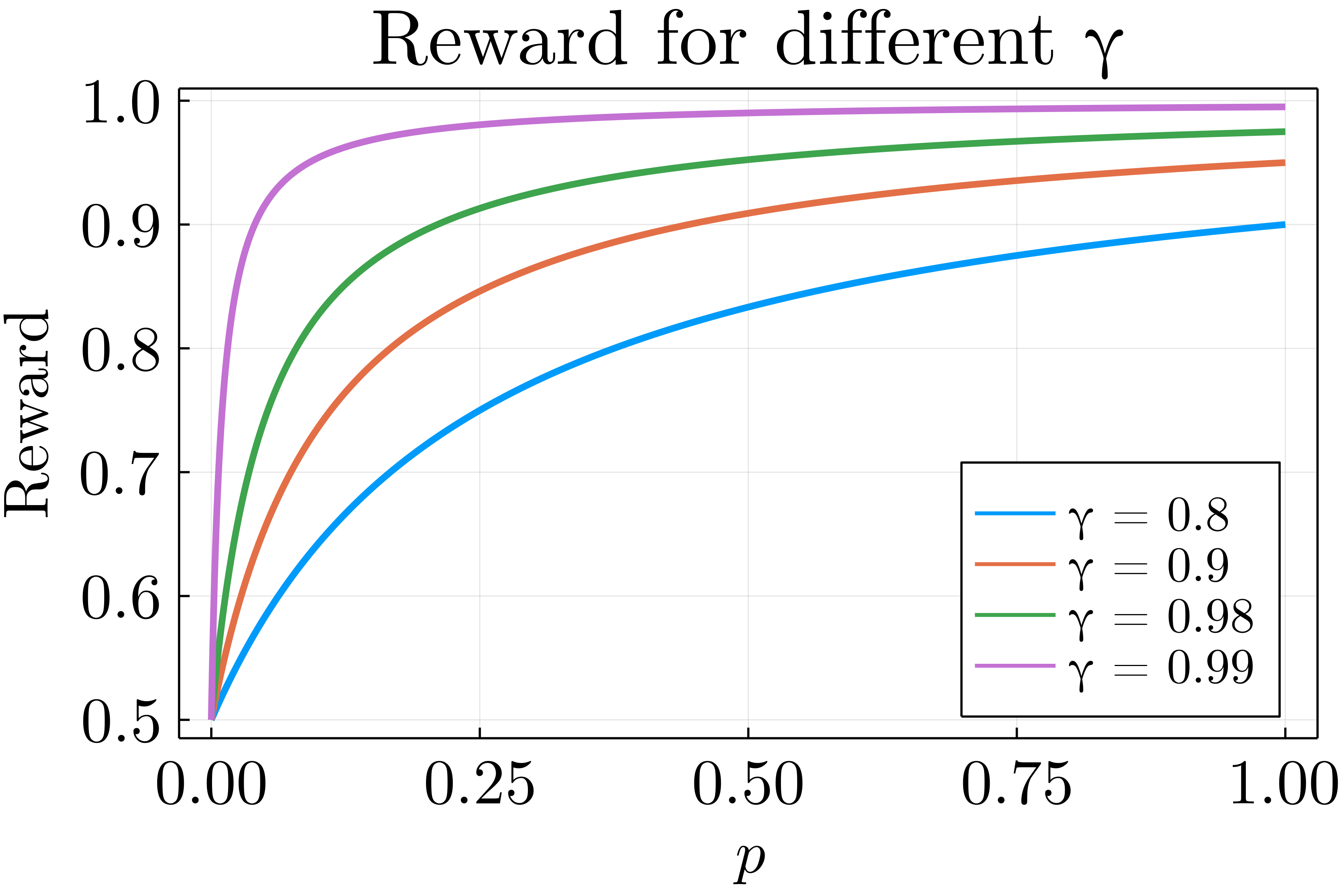}
    \includegraphics[width=0.18\textwidth]{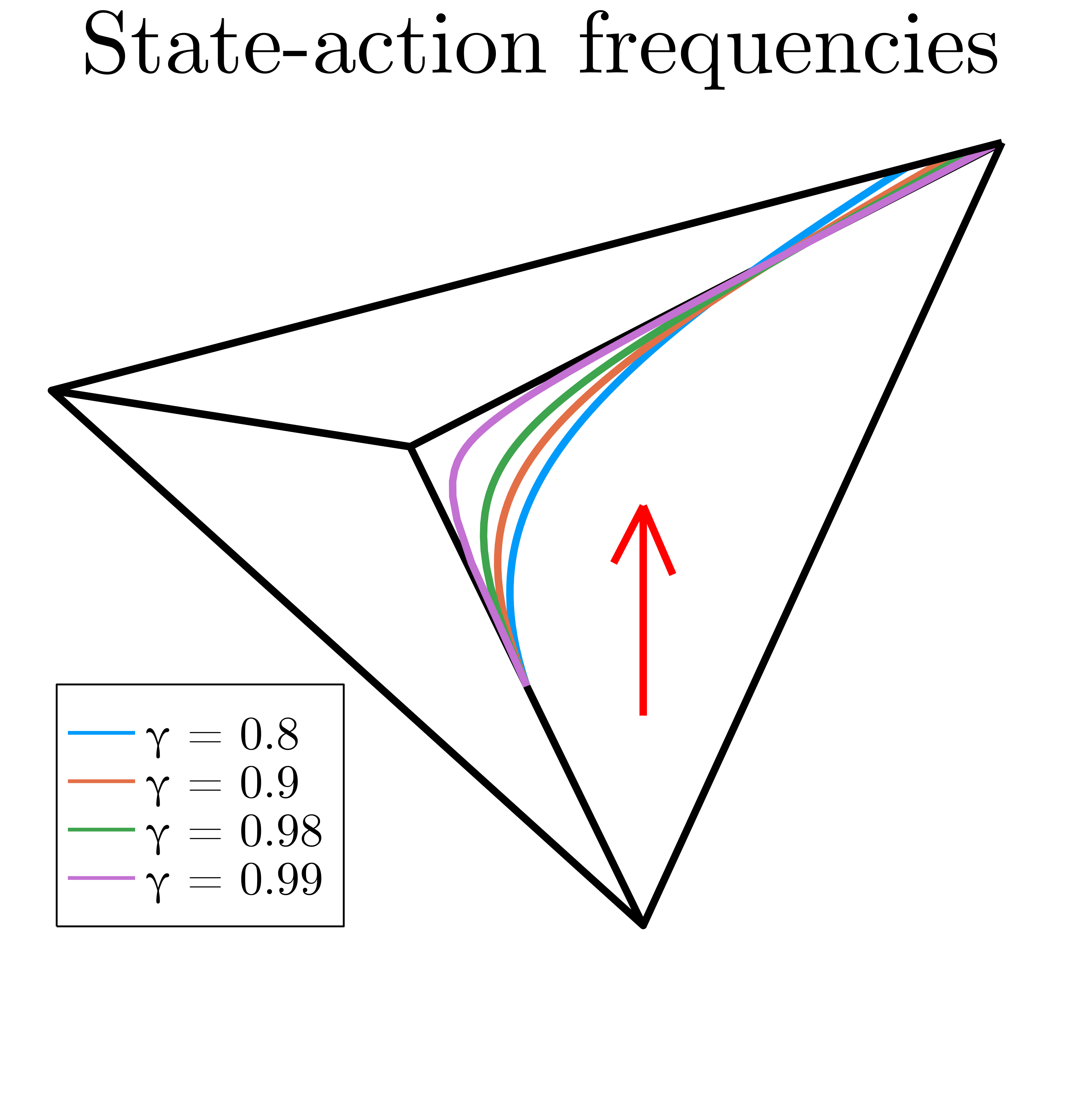}
    \vspace{-.5cm}
    \caption{
    A blind controller with two states; policies can be parameterized by $p\in[0, 1]
    $; for increasing 
    $\gamma$ the Lipschitz constant of the reward increases (left); 
    the corresponding feasible state-action frequencies, probability simplex $\Delta_{\SS\times\AA}$, and instantaneous reward vector (right). 
    }
    \label{fig:range4}
    \vspace{-.5cm}
\end{wrapfigure}

\paragraph{{Policy gradients}}
A very popular approach in Reinforcement Learning are policy gradients methods. 
In fully observable problems, the iteration complexity of policy gradient methods behaves like $O((1-\gamma)^{-\kappa})$, where $\gamma\in(0,1)$ is the discount factor and $\kappa\in\mathbb N$ depends on the specific method~\cite{cen2021fast}. 
This is reminiscent of the 
Lipschitz constant of the reward function (as a function of the policy), which behaves like $O((1-\gamma)^{-1})$~\cite{pirotta2015policy}; see Figure~\ref{fig:range4}. 
This leads to increasingly ill-conditioned problems as $\gamma\to 1$ and can cause undesired oscillations during optimization 
\cite{wagner2011reinterpretation}. 
However, choosing a discount factor close to $1$ is desirable 
as one often wishes to optimize the mean reward rather than a discounted reward. 
This is also required to prevent vanishing policy gradients 
in sparse reward MDPs, where, denoting $n_\SS$ the number of states, gradients can of order $O(2^{-n_\SS/2})$ if $\gamma \le n_\SS/(n_\SS +1)$~\cite{agarwal2021theory}. In principle the ill-conditioning problem can be addressed by introducing an appropriate metric, as in natural policy gradients or trust region policy optimization, which can be costly, however. 

\paragraph{{Optimization in state-action space}}

An alternative to optimizing over the policy parameters is to 
optimize the reward over all feasible state-action frequencies of the POMDP. 
{State-action frequencies are weighted averages of the time spent by the Markov process at different state-action pairs. The reward of a policy depends linearly on its state-action frequency and the optimization in state-action space maximizes the time spend in favorable states. 
The policy corresponding to a state-action frequency can be recovered by conditioning over states.} 
In MDPs, the state-action frequencies form a polytope and hence the problem becomes a linear program~\cite{derman1970finite,kallenberg1994survey}. 
This yields a strongly polynomial algorithmic approach, i.e., does not degrade for $\gamma\to  1$~\cite{post2015simplex}. 
In the case of POMDPs, additional polynomial constraints describe the set of all feasible state-action frequencies of POMDPs, which were recently described by~ \cite{mueller2021geometry}.
This yields a  
polynomial program of POMDPs generalizing the linear programs of MDPs. 
In this work we investigate the practical viability of this approach to optimize the reward in POMDPs. 
We consider navigation tasks in random mazes of different sizes, for which we develop a tool to 
generate the constraints and solve the constrained optimization problem using interior point methods. 
Our experiments show that the proposed method can yield significant computational savings compared to several baselines, while also remaining numerically stable across values of $\gamma$ where other methods fail. 

\section{Notation and setup}
We denote the simplex of probability distributions on a finite set \(\mathcal X\) by \(\Delta_{\mathcal X}\) and the set of Markov kernels from a finite set $\mathcal X$ to another finite set $\mathcal{Y}$ by \(\Delta_{\mathcal Y}^{\mathcal X}\). 
A \emph{partially observable Markov decision process} or shortly \emph{POMDP} is a tuple \((\mathcal S, \mathcal O, \mathcal A, \alpha, \beta, r)\). We assume that \(\mathcal S, \mathcal O\) and \(\mathcal A\) are finite sets which we call the \emph{state}, the \emph{observation} and the \emph{action space} respectively. We fix a Markov kernel \(\alpha\in\Delta_{\mathcal S}^{\mathcal S\times\mathcal A}\) which we call the \emph{transition mechanism} and a kernel \(\beta\in\Delta_{\mathcal O}^{\mathcal S}\) which we call the \emph{observation mechanism}. Further, we consider an \emph{instantaneous reward vector} \(r\in\mathbb R^{\mathcal S\times\mathcal A}\).  
As \emph{policies} we consider elements \(\pi\in\Delta_{\mathcal A}^{\mathcal O}\), which are referred to as \emph{memoryless stochastic policies}. Every policy defines a transition kernel \(P_\pi\in\Delta_{\mathcal S\times\mathcal A}^{\mathcal S\times\mathcal A}\) by $P_\pi(s^\prime, a^\prime|s, a) \coloneqq \alpha(s^\prime|s, a) \sum_o \pi(a^\prime|o)\beta(o|s^\prime)$.
For any initial state distribution \(\mu\in\Delta_{\mathcal S}\), a policy \(\pi\in\Delta_{\mathcal A}^{\mathcal O}\) defines a Markov process on \(\mathcal S\times \mathcal A\) with transition kernel \(P_\pi\) which we denote by \(\mathbb P^{\pi, \mu}\). For a \emph{discount rate} \(\gamma\in(0, 1)\)  
we define the \emph{infinite horizon expected discounted reward}
\begin{align}
    R(\pi) & \coloneqq \mathbb E_{\mathbb P^{\pi, \mu}}\left[ (1-\gamma) \sum_{t=0}^\infty\gamma^tr(s_t, a_t)\right].
    \end{align}
We consider the \emph{reward maximization problem}, i.e., the problem of maximizing $R(\pi)$ subject to $\pi\in\Delta_\AA^\OO$. 
{The discount factor $\gamma$ is most commonly introduced for mathematical convenience. For large state spaces and sparse rewards or more generally for POMDPs where the optimal policy should not depend on an unknown initial state distribution, it may be more desirable to consider the expected reward per time step, which corresponds to the limit $\gamma\to1$. }

\section{Feasible state-action frequencies}

It is well known that $R(\pi)= \langle r, \eta^\pi\rangle_{\SS\times\AA}$, where $\eta^\pi(s, a) = (1-\gamma) \sum_{t\ge0} \gamma^t \mathbb P^{\pi, \mu}(s_t=s, a_t=a)$
is the \emph{state-action frequency} of $\pi$. 
We denote the set of feasible state-action frequencies in the fully observable and in the partially observable case by $\mathcal N$ and $\mathcal N_\beta$, respectively. 
Hence, instead of solving the reward maximization problem over $\pi$, one can also solve the \emph{reward maximization problem in state-action space}, which is given by
\begin{equation}\label{eq:rewardMaximizationImplicit}
    \operatorname{maximize} \;\langle r , \eta \rangle \quad \text{subject to } \eta\in\mathcal N_\beta\subseteq\Delta_{\SS\times\AA}.
\end{equation}
To do this in practice, one requires a suitable characterization of the set of feasible state-action frequencies $
\mathcal N_\beta$.
To describe this set via polynomial conditions, we introduce {the \emph{effective policy poltyope\footnote{Here, $\pi\circ\beta$ denotes the composition of the Markov kernels given by $(\pi\circ\beta)(a|s)\coloneqq \sum_{o\in\OO}\pi(a|o)\beta(o|s)$. 
} $\Delta_\AA^{\SS,\beta}\coloneqq\{\pi\circ\beta:\pi\in\Delta_\AA^\OO\}$}}, {which is the set of state policies that can be realized by selecting the actions based on observations made according to $\beta$}. Note that the set of effective policies $\Delta_\AA^{\SS,\beta}$ is a polytope since it is the image of the polytope $\Delta_\AA^\OO$ under the linear map $\pi\mapsto\pi\circ\beta$. 

\paragraph{The state-action frequencies of an MDP}
As mentioned before, the state-action frequencies of an MDP form a polytope 
$  \mathcal N = 
    \left\{\eta\in\mathbb R_{\ge0} : 
    \ell_s(\eta) = 0
    \text{ for } s\in\SS \right\}
    \subseteq\Delta_{\SS\times\AA}
    $,
where $\ell_s(\eta) = \langle \delta_s\otimes\mathds{1}_\mathcal A - \gamma \alpha(s|\cdot, \cdot), \eta\rangle_{\SS\times\AA} - (1-\gamma)\mu_s$; see~\cite{derman1970finite}. 
Hence for MDPs the reward maximization problem~\eqref{eq:rewardMaximizationImplicit} becomes a linear program in state-action frequency space. This is known as the dual linear programming formulation of MDPs~\cite{kallenberg1994survey}.

\paragraph{The state-action frequencies of a POMDP}
{The effective policy corresponding to a state-action frequency can be computed by conditioning. 
In order for the conditioning to be well defined}, we require the following assumption, which holds for instance if the initial distribution $\mu$ has full support. 
\begin{assumption} 
\label{ass:positivity}
For any state-action frequency $\eta\in\mathcal N$ and any state $s\in\SS$ it holds that $\sum_a \eta_{sa}>0$. 
In the mean reward case we further assume that for every policy $\pi\in\Delta_\AA^\OO$ there exists a unique stationary distribution of $P_\pi$. This is a standard assumption in MDP linear programming~\cite{kallenberg1994survey} and necessary for the convergence of PG methods~\cite{mei2020global}. 
\end{assumption}

{The correspondence of state-action frequencies $\eta\in\mathcal N$ and state policies $\tau\in\Delta_\AA^\SS$ via conditioning provides a correspondence of polynomial inequalities in the two sets $\Delta_\AA^\SS$ and $\mathcal N$.
}
More precisely, setting $S\coloneqq \{s\in\mathcal S : b_{sa} \ne0 \text{ for some } a\in\mathcal A\}$ it holds that 
\begin{equation}\label{eq:correspondence}
    \sum_{s, a} b_{sa} \tau_{sa} \ge 0 \quad \text{if and only if} \quad \sum_{s\in S} \sum_a b_{sa}\eta_{sa}\prod_{s'\in S\setminus\{s\}}\sum_{a'} \eta_{s'a'}\ge0.
\end{equation}
{Therefore, the set of feasible state-action frequencies $\mathcal N_\beta$ can be described by finitely many polynomial (in)equalities corresponding to the linear (in)equalities describing $\Delta_\AA^{\SS,\beta}$ in $\Delta_\AA^\SS$. In particular, this shows that solving the reward optimization problem in infinite-horizon POMDPs with memoryless stochastic policies is equivalent to a polynomially constrained optimization problem with linear objective, which generalizes the lienar program associated to MDPs. This formulation was obtained in \cite{mueller2021geometry} and was used to establish upper bounds on the number of critical points of the reward optimization problem. In this work, we focus the solution of POMDPs via this polynomial program.} 

\section{Reward optimization in state-action frequency space (ROSA)}

We formulate our approach for 
\textbf{R}eward \textbf{O}ptimization in \textbf{S}tate-\textbf{A}ction space (ROSA) in Algorithm~\ref{alg:POSA}. 
\begin{algorithm}
\caption{
\textbf{R}eward \textbf{O}ptimization in \textbf{S}tate-\textbf{A}ction space (ROSA)}\label{alg:POSA}
\begin{algorithmic}[1]
\Require 
$\alpha\in\Delta_\SS^{\SS\times\AA}$, 
$\beta\in\Delta_\OO^\SS$, 
$\gamma\in (0, 1)$, 
$\mu\in\Delta_\SS$
\ForAll{$s\in\mathcal S$}
    \State $\ell_s(\eta) \gets \langle \delta_s\otimes\mathds{1}_\mathcal A - \gamma \alpha(s|\cdot, \cdot), \eta\rangle_{\SS\times\AA} - (1-\gamma)\mu_s$ \Comment{Define the linear equalities}
\EndFor
\State Compute the defining linear inequalities of $\Delta_\AA^{\SS,\beta}$ \Comment{In closed form or algorithmically} \label{line:definingLinear}
\State Compute the defining polynomial inequalities $p_i(\eta)\ge0$ of $\mathcal N_\beta$ \Comment{Can be done according to~\eqref{eq:correspondence}}
\State $\eta^\ast\gets\arg\max\langle r, \eta \rangle$ sbj to $\eta\ge0$, $\ell_s(\eta) = 0$
, $p_{i}(\eta)\ge0$ \Comment{Solve the constrained maximization problem} \label{line:solve}
\State $R^\ast\gets \langle r, \eta^\ast \rangle$ \Comment{Evaluate the optimal value}
\State $\tau^\ast\gets \eta^\ast(\cdot|\cdot)\in\Delta_\mathcal A^\mathcal S$ \Comment{Compute an optimal state policy}
\State $\pi^\ast\gets$ solution of $\beta\pi = \tau^{\ast}$ \Comment{Compute an optimal observation policy}

\Return $\eta^\ast$, $R^\ast$, $\pi^\ast$ \Comment{maximizer, optimal value, optimal policy}
\end{algorithmic}
\end{algorithm}
The two non-trivial steps in the algorithm are the computation of the defining linear inequalities of the polytope $\Delta_\AA^{\SS,\beta}$ in line~\ref{line:definingLinear} and the solution of the constrained optimization problem in line~\ref{line:solve}. 

\paragraph{Computing the polynomial constraints}
The linear inequalities defining $\Delta_\AA^{\SS,\beta}$ and therefore the polynomial constraints of $\mathcal N_\beta$ can be computed in closed form if $\beta$ has linear independent columns~\cite[Thm.\ 12]{mueller2021geometry}. If this is not the case, they can be computed algorithmically using Fourier-Motzkin elimination, block elimination, vertex approaches, or equality set projection~\cite{Jones:169768}. 
Let us discuss the special case of deterministic observations. We associated $\beta$ with a mapping $\SS\to\OO$, which partitions the state space into sets $S_o \coloneqq \{s\in\SS : \beta(o|s) = 1\}\subseteq\mathcal S$. 
If we fix an arbitrary action $a_0\in\AA$ and arbitrary states $s_o\in S_o,o\in\mathcal{O}$, the polynomial equations cutting out the set $\mathcal N_\beta$ of feasible state-action frequencies from the set $\mathcal N$ of all state-action frequencies are given by
\begin{equation}
    p^o_{sa}(\eta) \coloneqq \eta_{s_oa} \sum_{a'} \eta_{sa'} - \eta_{sa} \sum_{a'} \eta_{s_oa'} = \sum_{a'\ne a} (\eta_{s_oa}\eta_{sa'} - \eta_{s_oa'} \eta_{sa}) = 0, 
\end{equation}
for all actions $a\in\AA\setminus\{a_0\}$, states $s\in S_o\setminus\{s_o\}$ and observations $o\in\OO$. 
Hence, for deterministic observations the reward maximization problem in state-action space takes the form
\begin{equation}\label{eq:polyFormRewMax}
    \operatorname{maximize} \;\langle r, \eta \rangle \quad \text{subject to } 
    \left\{
\begin{array}{rl}
    \ell_s(\eta) = 0 & \text{for } s\in\mathcal S\\
    p_{sa}^o(\eta) = 0 & \text{for } o\in \OO, a\in\AA\setminus\{\tilde a\}, s\in S_o\setminus\{s_o\} \\
    \eta_{sa} \ge 0 & \text{for } s\in\SS, a\in\AA.
\end{array}
\right.
\end{equation}
This is a problem in $\lvert\SS\rvert\lvert\AA\rvert$ variables with $\lvert\SS\rvert$ linear and $\sum_{o}(\lvert S_o\rvert - 1) (\lvert \AA\rvert -1)=(\lvert \SS\rvert - \lvert \OO\rvert)(\lvert \AA\rvert-1)$ quadratic equality constraints and $\lvert\SS\rvert\lvert\AA\rvert$ inequality constraints (of which only $\lvert \OO\rvert\lvert\AA\rvert$ are non redundant).

\paragraph{Implementation and solution of the optimization problem}
We provide a \texttt{Julia}~\cite{bezanson2017julia} implementation of ROSA for deterministic observations. 
In general, problem~\eqref{eq:polyFormRewMax} can be solved with any constrainted optimization solver. 
Our implementation is built on \texttt{Ipopt}, an interior point line search method~\cite{wachter2006implementation}. 
We call \texttt{Ipopt} via the modeling language \texttt{JuMP} in which the constraints are easy to implement~\cite{dunning2017jump}. 
The implementation is available under~\url{https://github.com/muellerjohannes/POMDPs-ROSA}.

\section{Experiments}
\label{sec:experiments}

To demonstrate the performance of ROSA we test it on navigation problems in mazes. For this, we generate connected mazes using a random depth first search~\cite{mazeGeneration}. 
Then we randomly select a state as the goal state at which a reward of $\lvert\SS\rvert$ is picked up and from which the agent transitions to a uniform state. For all other states four actions move the agent right, left, up or down. The agent can only observe the 8 neighboring cells and starts at a uniform position. 

We compare against two other optimization approaches. 
First, we consider tabular softmax policies and directly optimize the parameters for the exact discounted reward. 
Instead of a vanilla policy gradient ascent, we use L-BFGS, which is a first order method that estimates second order information.
In comparison to a naive policy gradient, we observed L-BFGS to converge faster. 
We refer to this approach as direct policy optimization (DPO). 
As a second baseline we consider the reformulation of the reward maximization problem as a quadratically constrained linear program~\cite{amato2006solving}
\begin{equation}\label{BCP}
    \operatorname{maximize} \;\langle \mu, v \rangle \quad \text{subject to }
    \pi\in\Delta_\AA^\OO \text{ and }
    v = \gamma p_\pi v + (1-\gamma) r_\pi
    ,
\end{equation}
where $p_\pi(s'|s) \coloneqq \sum_{o,a}\pi(a|o)\beta(o|s) \alpha(s'|s,a)$ and $r_\pi(s)\coloneqq \sum_{a,o} r(s,a)\pi(a|o)\beta(o|s)$.

Note that here the constraint is on the value function. We use \texttt{Ipopt} to solve~\eqref{BCP}. We call this approach \emph{Bellman constrained programming} (BCP). 

\begin{figure}[h!] 
    \centering
\begin{tikzpicture}
\node at (0,0) {
    \includegraphics[width=0.246\textwidth]{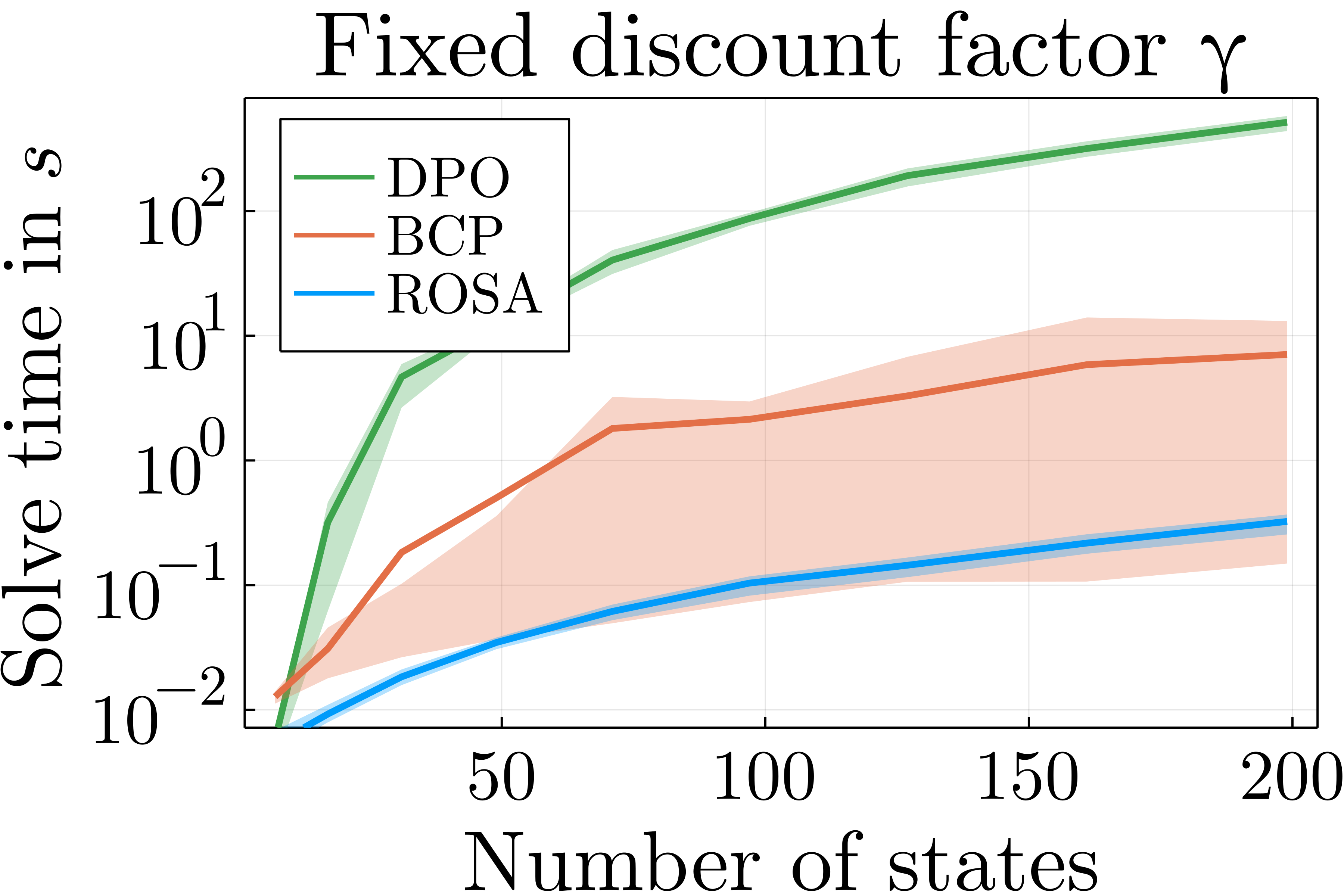} 
    \includegraphics[width=0.246\textwidth]{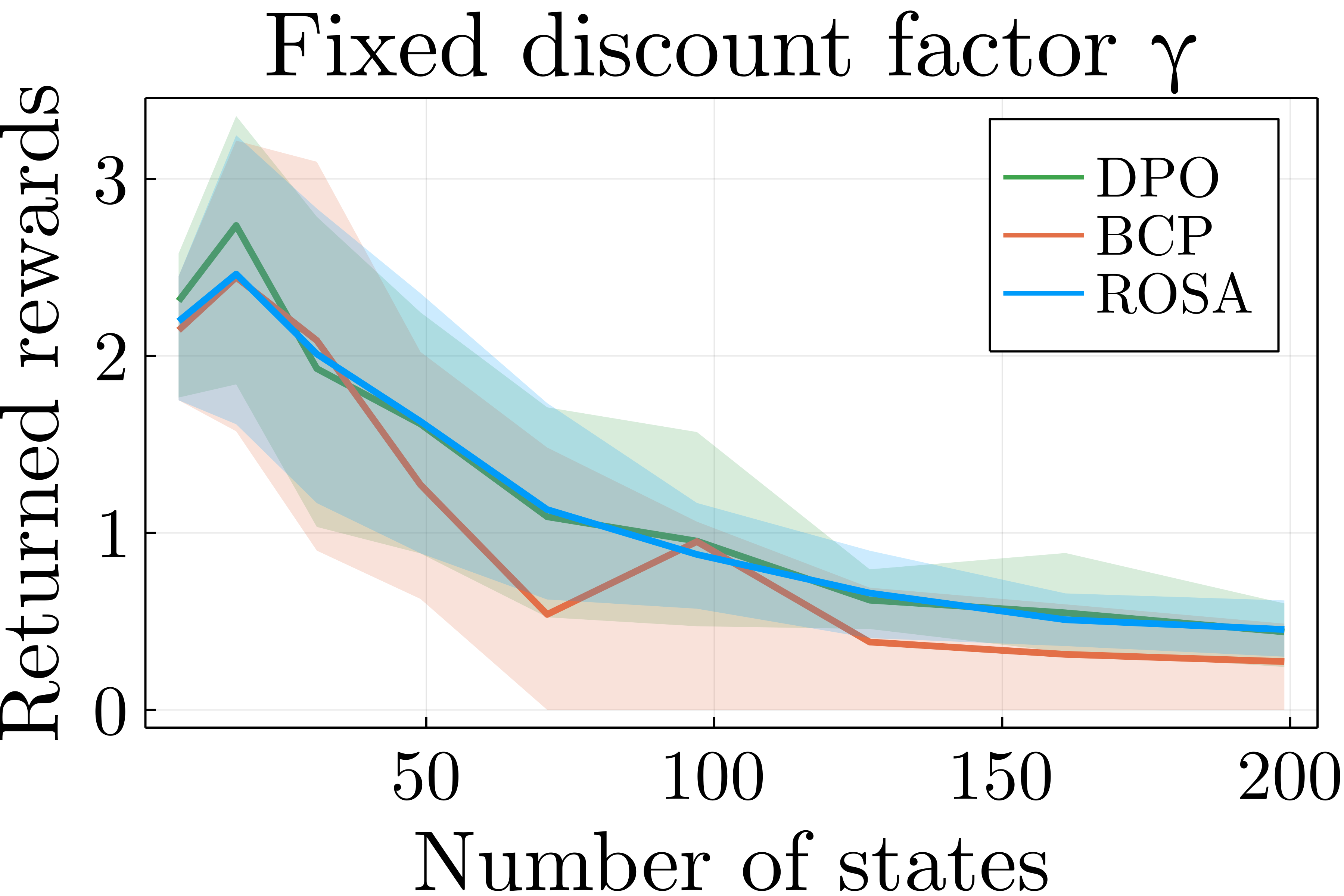} 
    \includegraphics[width=0.246\textwidth]{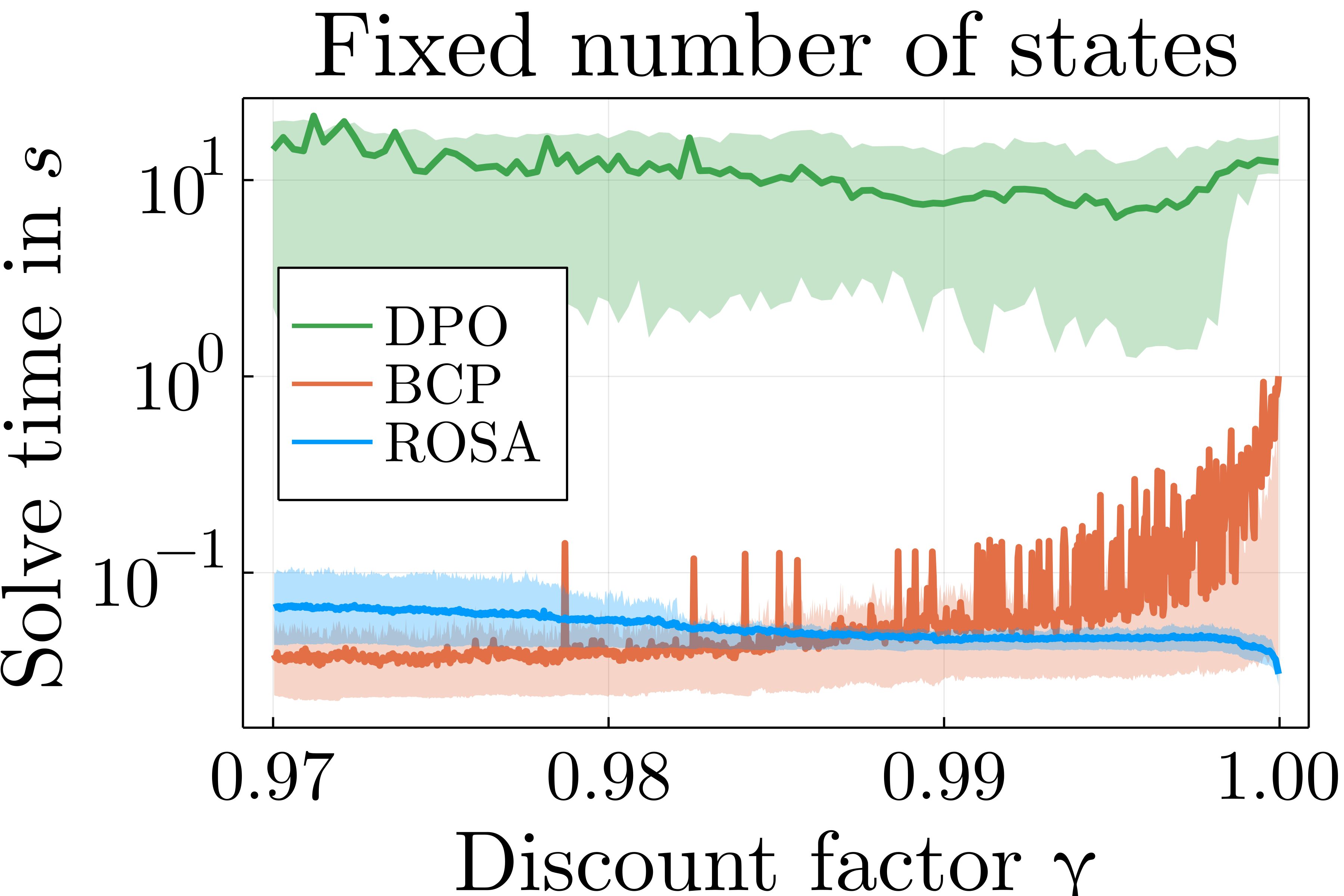} 
    \includegraphics[width=0.246\textwidth]{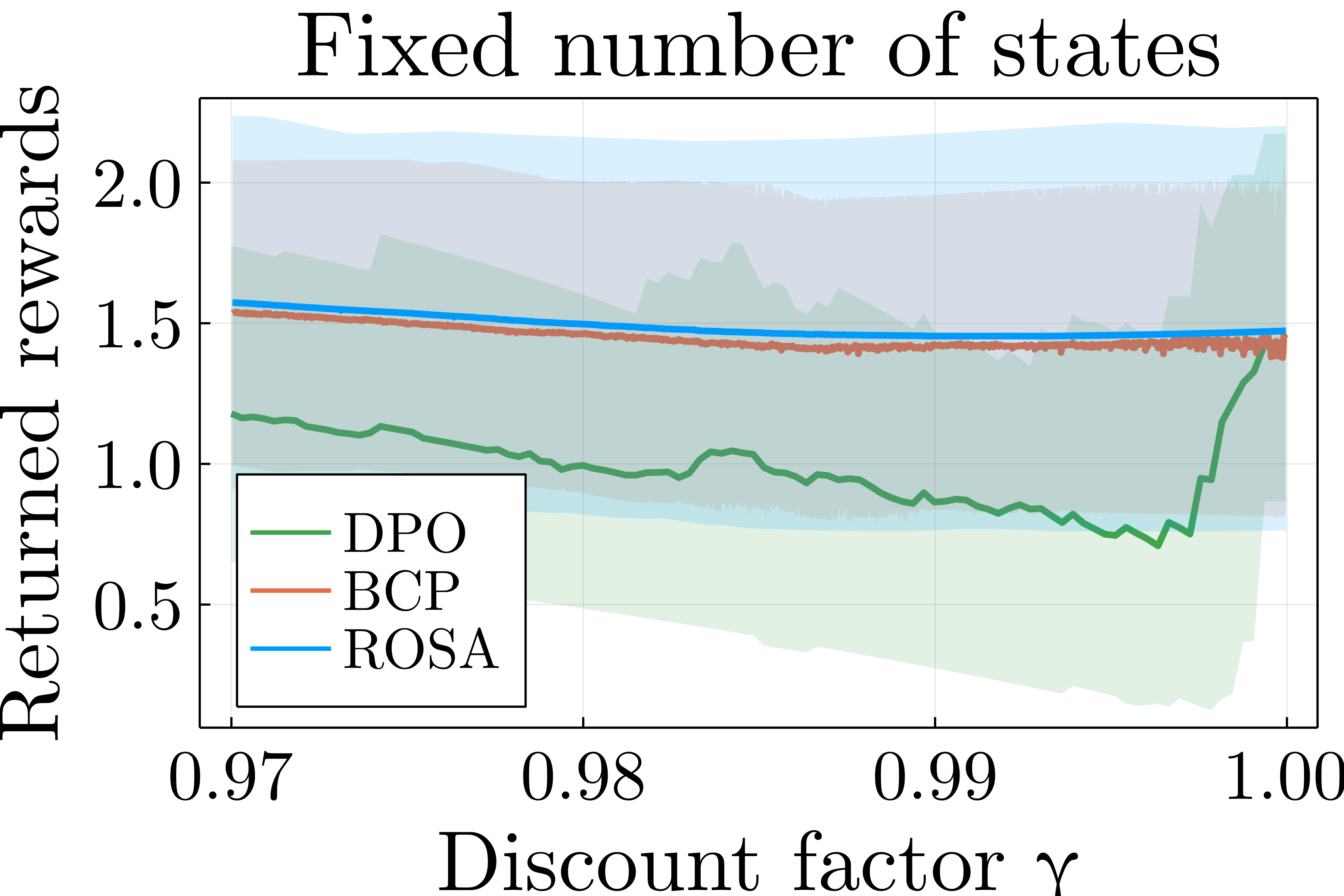}
};
\node at (-1.9,.6) {    
    \includegraphics[width=0.05\textwidth]{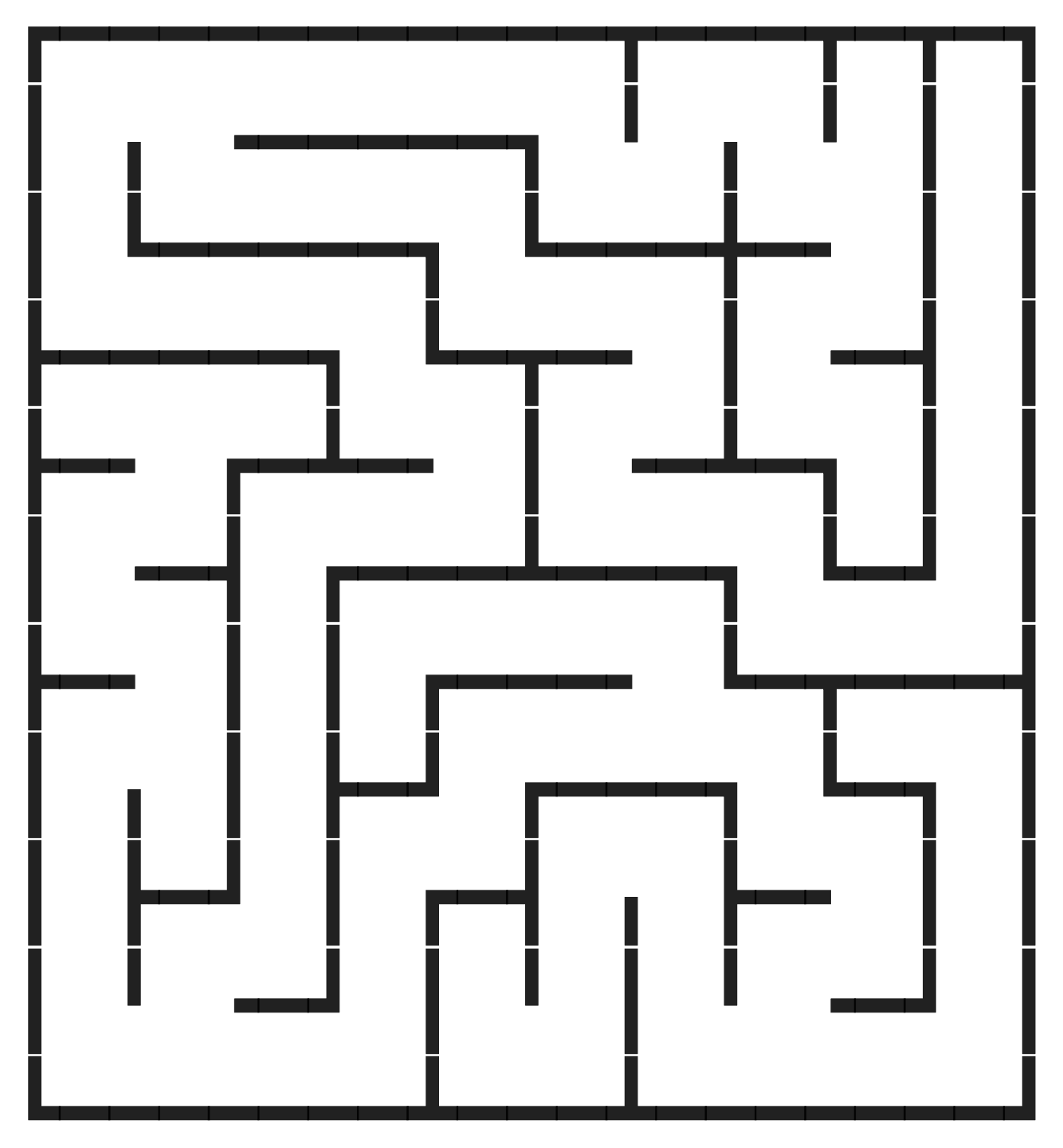}};
\end{tikzpicture}
\vspace{-.5cm}
    \caption{Shown is the solution time and cumulative reward obtained by different methods solving navigation tasks depending on the number of states or the discount factor. Inset shows one of the mazes with $199$ states. ROSA reaches higher reward in less time, with stability improvements and time savings becoming more pronounced for larger $\gamma$. }
    \label{fig:plots}
\end{figure}

In order to compare the running times of the three approaches, we generate square mazes of side length $2n-1$ and $2n^2-1$ states, for $n = 2, \dots, 10$. 
We solve the POMDPs for a discount factor of $\gamma = 0.9999$ using ROSA, BCP and DPO for $10^2$ different mazes of each size\footnote{For DPO we solved only $20$ mazes of each size due to the long solution time.} and report the mean solution times and achieved rewards as well as their $16\%$ and $84\%$ quantiles in Figure~\ref{fig:plots}. 
We observe that all three methods achieve comparable rewards. However, DPO becomes inefficient even for problems of moderate size and the running time of BCP grows significantly faster compared to ROSA. 

To evaluate the performance of ROSA for $\gamma\to1$ we solve $10^2$ mazes\footnote{For DPO we consider only 30 mazes and $10^2$ values of $\gamma$.} with side length $9$ and $49$ states for increasing discount factors. We report the average solution times and achieved reward in Figure~\ref{fig:plots}.
In the comparison of the rewards, examples where BCP did not converge are excluded. In these experiments we see that BCP becomes unstable, whereas the solution time of ROSA {appears to be very robust and even decrease} for $\gamma\to1$. 
In fact, in the solution of~\eqref{BCP} \texttt{Ipopt} fails to converge to local optimality for about $15\%$ of all problems with discount factor at least $0.9999$. 

\bibliographystyle{amsalpha}
\bibliography{bib}

\end{document}